\documentclass[AMA,STIX1COL]{WileyNJD-v2}

\articletype{Article Type}%

\usepackage{arydshln}
\usepackage{graphicx}
\usepackage{multirow}
\usepackage{array}
\usepackage{makecell}
\usepackage{float}
\usepackage{booktabs}
\usepackage{xcolor}
\usepackage{subcaption}
\usepackage{mwe}
\usepackage{epsfig}
\usepackage{xr}
\usepackage{endfloat} 

\newcommand{\boldbeta}{\mbox{\boldmath $\beta$}}

\newcommand{\boldtheta}{\mbox{\boldmath $\theta$}}
\newcommand{\boldx}{\mbox{\boldmath $x$}}

\newcommand{\boldw}{\mbox{\boldmath $w$}}

\begin{document}

\title{Deep Neural Network-based Accelerated Failure Time Models Using Rank Loss}
    \author[1,4]{Gwangsu Kim}
    \author[3]{Jeongho Park}
    \author[2,3]{Sangwook Kang}
    \authormark{KIM \textsc{et al}}
    \address[1]{\orgdiv{Department of Statistics (Institute of Applied Statistics)}, \orgname{Jeonbuk National University}, \orgaddress{\state{Jeonju}, \country{Republic of Korea}}}
    \address[2]{\orgdiv{Department of Applied Statistics}, \orgname{Yonsei University}, \orgaddress{\state{Seoul}, \country{Republic of Korea}}}
    \address[3]{\orgdiv{Department of Statistics and Data Science}, \orgname{Yonsei University}, \orgaddress{\state{Seoul}, \country{Republic of Korea}}}
   \address[4]{\orgdiv{Center for Advanced Image and Information Technology}, \orgname{Jeonbuk National University}, \orgaddress{\state{Jeonju}, \country{Republic of Korea}}}
    \corres{*Sangwook Kang, Department of Applied Statistics / Department of Statistics and Data Science, Yonsei University, Seoul, Korea. \email{kanggi1@yonsei.ac.kr}}
    \corres{*Sangwook Kang, Department of Applied Statistics / Department of Statistics and Data Science, Yonsei University, Seoul, Korea. \email{kanggi1@yonsei.ac.kr}
    }




\abstract[Summary]{An accelerated failure time (AFT) model assumes a log-linear relationship between failure times and a set of covariates. In contrast to other popular survival models that work on hazard functions, the effects of covariates are directly on failure times, the interpretation of which is intuitive. The semiparametric AFT model that does not specify the error distribution is sufficiently flexible and robust to depart from the distributional assumption. Owing to its desirable features, this class of model has been considered a promising alternative to the popular Cox model in the analysis of censored failure time data. However, in these AFT models, a linear predictor for the mean is typically assumed. Little research has addressed the non-linearity of predictors when modeling the mean. Deep neural networks (DNNs) have received much attention over the past few decades and have achieved remarkable success in a variety of fields. DNNs have a number of notable advantages and have been shown to be particularly useful in addressing non-linearity. Here, we propose applying a DNN to fit AFT models using Gehan-type loss combined with a sub-sampling technique. Finite sample properties of the proposed DNN and rank-based AFT model (DeepR-AFT) were investigated via an extensive simulation study. The DeepR-AFT model showed superior performance over its parametric and semiparametric counterparts when the predictor was non-linear. 
For linear predictors, DeepR-AFT performed better when the dimensions of the covariates were large. The superior performance of the proposed DeepR-AFT was demonstrated using three real datasets.}

\keywords{C-index, Gehan loss, Non-linear mean function, Semiparametric accelerated failure time model, Survival analysis}

\maketitle


\section{Introduction} \label{sec:int}
    For regression modeling of failure times, an accelerated failure time (AFT) assumption is often postulated. The AFT model relates a log-transformed failure time to a linear combination of regression coefficients and a set of covariates added by a zero-mean random error term.  
    The AFT model can be either parametric or semiparametric, depending on whether the distribution of the error term is assumed to be parametric or is left unspecified, respectively. The latter shares the semi-parametric property of the well-known Cox proportional hazards model \citep{Cox:regr:1972} and, as such, is viewed as a viable alternative.
    Parametric AFT (PAFT) models can be inferred using the usual maximum likelihood method. For semiparametric AFT (SAFT) models, rank-based methods have been popular \citep{Pren:line:1978,Jin:Lin:Wei:Ying:rank:2003,Jin:Lin:Ying:rank:2006,John:Stra:indu:2009} among others, including the least-squares estimation \citep{Buck:Jame:line:1979,stute1993consistent,Jin:Lin:Ying:on:2006}. 
    The theoretical properties of the rank-based estimators have been investigated rigorously and are now well established \citep{Tsia:esti:1990, Ying:larg:1993, Fyge:Rito:mono:1994}. 
    Popular statistical software such as R implements some of these methods, which have made fitting censored survival data based on AFT models routine analysis. For example, the maximum likelihood estimators (MLEs) for regression coefficients in PAFT models can be obtained by 
    the \texttt{survreg} function in the \texttt{survival} package in \texttt{R} \citep{survival-package}. Rank-based estimators for SAFT models are implemented in the \texttt{aftgee} package in \texttt{R} \citep{aftgee-jss}. 
    
 The aforementioned development of the AFT modeling approach is based mostly on the assumption that the effects of covariates are linear or, more generally, parametric. This simplified depiction of the covariate effect has several benefits, including a direct summary of the covariate effect that allows its straightforward and intuitive interpretation. 
Meanwhile, when the underlying relationship between covariates and failure times is complex and non-linear, this linear modeling approach may be too restrictive and lead to biased results, as it does not properly reflect the true underlying relationship. This could be even more problematic when predicting survival times is the main interest, as is the case in many clinical and biological studies. In this situation, a more accurate prediction is anticipated by imposing a non-linear relationship that nonparametrically models the mean function part. Some efforts have been made to allow for nonparametric modeling of AFT models under the framework of the smoothing spline with penalization \citep{leng2007accelerated}. Although interaction terms can also be handled in this approach, low-order terms are typically considered to control the model complexity 
\citep{huang2000functional,leng2007accelerated}. Deep neural networks (DNNs) have been shown to be particularly successful in addressing non-linearity and interactions \citep{katzman2018deepsurv,kvamme2019time}. In this manuscript, we propose a DNN-based model and algorithm referred to as DeepR-AFT. 
    
Considering the recent success of DNNs, applying a DNN to AFT models can be an effective method for addressing non-linearity. By virtue of their model complexity and optimization techniques, DNNs have been applied successfully in survival analysis prediction, as reported in a variety of fields, including medicine \citep{he2020deep,chattopadhyay2020predicting, lundervold2019overview}. The DNN's capacity to model and predict nonparametric functions with large dimensional variables and their interactions is one of its many prominent advantages. DNNs are known for having strong representation power and have been adapted to numerous tasks of natural language processing and reinforcement learning.
Recent efforts have been made to use DNNs in the context of survival analysis, which commonly uses Cox or discretized hazard models \citep{katzman2018deepsurv, lee2018deephit, chapfuwa2018adversarial, kvamme2019time, zhao2020deep, rava2020deephazard}. A DNN-cooperative meta-learning algorithm was proposed more recently \citep{qiu2020meta} that demonstrated remarkable performance in the use of small datasets through a “learn-to-learn” approach. 
Except for Kvamme {\it et al.}\cite{kvamme2019time}, who proposed the use of mini-batches in DNNs, most previous works based on a partial-likelihood structure use a full batch in the stochastic gradient (SGD) algorithm. This is mainly due to the existence of risk sets, a stochastic property in the popular partial likelihood. In contrast, the Gehan loss 
{\rm for the AFT models does not depend on risk sets while taking into account all pairs.} 
With this property, the application of the mini-batch algorithm becomes easily feasible and, thus, leads to achieving scalability. The original computational order is, however, quadratic in sample size. {\rm To alleviate this computational burden}, we propose a sub-sampling procedure that can reduce the computational order to linear in the scale of the sample size. Since the resulting sample can be regarded as a random sample from all possible pairs, the SGD in mini-batches can be readily applied. The existing literature rarely addresses this scalability using a DNN in survival analysis.
Additionally, limited research has been conducted on the use of DNNs with AFT models. The deepAFT\citep{chen2019} uses a DNN in the SAFT model, primarily via Buckley-James or inverse-probability-censoring-weight type losses. Surprisingly, the Gehan loss, a popular loss for SAFT models, has rarely been applied with a DNN in the published literature.
In addition, a deep extended hazard model \citep{zhong2021deep} has been proposed that considers general models, including the AFT model as a special case. The loss function is, therefore, not specific to the AFT model. When an AFT model is considered, it can diminish the accuracy of predictions compared to the Gehan loss. In turn, we propose to study a more specific DNN that is suitable for an AFT model, using a Gehan-type loss combined with a sub-sampling technique.

The rest of the paper is organized as follows. In Section~\ref{sec:mod}, a summary of existing approaches for fitting AFT models is provided. The proposed DNN modeling approach for non-linear AFT models is introduced in Section~\ref{sec3:dnn}. 
Section~\ref{sec:sim} presents the results of the extensive simulation experiments conducted under various aspects, including different mean functions, bias and variance trade-offs, and large dimensional covariates. We illustrate our proposed DNN approach using three real datasets in Section~\ref{sec:data}. In Section~\ref{sec:disc}, we conclude with a summary and a discussion of possible future research directions.   


\section{Conventional parametric and semiparametric AFT modeling}  \label{sec:mod}
\subsection{Setup and model} \label{sec:mod:setmod}
Let $T$ denote the potential failure time. $T$ may not be observable due to right censoring. Let $C$ denote the potential censoring time. Then, the observed time is $Y = \min(T, C)$ and $\Delta = \mathbb{I}(T \leq C)$ denotes the indicator for failure where the $\mathbb{I}(\cdot)$ is an indicator function. We assume that $T$ and $C$ are conditionally independent given $\boldx$, where $\boldx$ is a $p$-dimensional vector of covariates. Data for $n$ subjects can be represented by independent and identically distributed ($i.i.d$) copies of $(Y, \Delta, x)$, $(Y_i, \Delta_i, \boldx_i), i=1, \ldots, n$. 
We impose an AFT model for $T_i$ given $\boldx_i$, such as 
\begin{align}
    \log T_i = f(\boldx_i) + \epsilon_i, 
        \ i=1, \ldots n,\label{aft_model1}
\end{align}
where $f(\boldx_i)$ can be set to a linear or non-linear function  
and $\epsilon_i$ is a zero-mean random error term with a finite variance. 

\subsection{Inference for PAFT models} \label{sec:mod:paft}

When the distribution of $\epsilon_i$ is known,~(\ref{aft_model1}) is referred to as a PAFT model. 
We typically consider the linear case for the mean function, i.e., $f(\boldx_i) = \boldx_i^{\top} \boldbeta$.
PAFT models are often represented by $\log T_i = \boldx_i^{\top} \boldbeta + \sigma \epsilon_i$, 
where $\sigma$ is a scale parameter.
For inferences, likelihood-based approaches have typically been employed. The likelihood function under the setup described in Section~\ref{sec:mod:setmod} is given by 
\begin{equation*}
    L(\boldtheta) = \prod_{i=1}^{n} f_{T}(y_i; \boldtheta)^{\Delta_i}S_{T}(y_i; \boldtheta)^{1-\Delta_i}     
\end{equation*}
where $f_{T}(\cdot; \boldtheta)$ and $S_{T}(\cdot; \boldtheta)$ are the probability density function and survival function for $T$ that represent a contribution of an uncensored failure time and censored failure time, respectively. $\boldtheta = (\boldbeta^{\top}, \sigma)^{\top}$. 
The estimator of $\boldtheta$ is then defined as the maximizer of $L(\boldtheta)$, i.e., the MLE of $\boldtheta$. 
Its large sample properties, such as consistency and asymptotic normality, are well established under certain regularity conditions
\citep{klein2006survival, collett2015modelling}, based on wide-studied ML theories.   

\subsection{Rank-based inference for SAFT models} 
\label{sec:mod:saft}
Unlike PAFT models, in SAFT models, the distribution of $\epsilon_i$ is left unspecified. 
We still consider $f(\boldx_i) = \boldx_i^{\top} \boldbeta$ as the linear relationship for the mean function.
Rank-based estimators for SAFT models are based on weighted estimating functions constructed using the ranks of the censored residuals, $e_i(\boldbeta) = \log Y_i - \boldx_i^{\top}\boldbeta, i=1, \ldots, n$. 
This approach aligns conceptually with the widely utilized nonparametric test that employs rank sums.\cite{fygenson1994monotone}
Specifically, estimating functions with the Gehan-type weight is given by \citep{Pren:line:1978,Jin:Lin:Wei:Ying:rank:2003}
\begin{equation} \label{wee_rank_geh}
    U_{G}(\boldbeta ) = \sum_{i=1}^n \sum_{j=1}^n \Delta_i (\boldx_i - \boldx_j) \mathbb{I} \left\{e_j(\boldbeta ) \geq e_i(\boldbeta ) \right\}. 
\end{equation}
The rank-based estimator for $\boldbeta$ is then defined as the solution to $U_{G}(\boldbeta ) = 0$.
Note that (\ref{wee_rank_geh}) is the gradient of the convex function 
\begin{eqnarray}
    \sum_{i=1}^n \sum_{j=1}^n \Delta_i \left[ e_i ( \boldbeta) - e_j ( \boldbeta)   \right]^{-},
 \label{eqn:g1loss}
\end{eqnarray}
where $[a]^{-}$ denotes $| \min ( a, 0) |$. Therefore, the rank-based estimator for $\boldbeta$ can be equivalently obtained by minimizing \eqref{eqn:g1loss}. The resulting estimator for $\boldbeta$ has been shown to be consistent and asymptotically normally distributed \citep{Rito:esti:1990,Ying:larg:1993,Jin:Lin:Wei:Ying:rank:2003}. An induced smoothed version of \eqref{wee_rank_geh} is a more recently proposed variant of rank-based estimators and has the following form:
 \begin{equation*} \label{wee_rank_geh_ind}
     \tilde{U}_{G}(\boldbeta) = \sum_{i=1}^n \sum_{j=1}^n \Delta_i (\boldx_i -  \boldx_j)\Phi\left\{\frac{e_j(\boldbeta) - e_i(\boldbeta)}{r_{ij}} \right\},
 \end{equation*}
where $r_{ij} = n^{-1}(\boldx_i - \boldx_j)^{\top}(\boldx_i - \boldx_j)$ and $\Phi(\cdot)$ is the cumulative density function for the standard normal random variable. This modified rank-based estimator has been shown to be asymptotically equivalent to the estimator solving $U_G(\boldbeta) = 0$ or minimizing \eqref{eqn:g1loss} while providing a computationally more efficient variance estimate \citep{Brow:Wang:indu:2007,John:Stra:indu:2009,Chio:Kang:Yan:semi:2014}.
 
\section{DNN modeling in the non-linear-nonparametric AFT model} 
\label{sec3:dnn}
\subsection{Non-linear-nonparametric AFT model}
When $f$ is non-linear, and the error distribution is not specified, it is referred to as a non-linear-nonparametric AFT model. Here, we adopt a parameter ${\bf w}$ to model the non-linear $f$ without transforming covariates. For example, spline basis functions can be used for $f_{\bf w}$ to model the non-linearity. 
We propose to model $f$ using DNN to address effectively the possible non-linearity with a complex structure, such as interactions. To infer a non-linear-nonparametric AFT model, we also propose to use the following Gehan loss:
\begin{eqnarray}
    \sum_{i=1}^n \sum_{j=1}^n \Delta_i \left[ e_i ( {\bf w} ) - e_j ( {\bf w})   \right]^{-},
\label{eqn:g2loss}
\end{eqnarray}
where $e_i({\bf w}) = \log Y_i - f_{\bf w} ( {\boldsymbol x}_i) , i=1, \ldots, n$. 
\subsection{DeepR-AFT}
The loss function~(\ref{eqn:g2loss}) is designed to extend the mean modeling from ${\boldsymbol x}^{\top} \boldbeta$ to non-linear-nonparametric $f_{\bf w}({\boldsymbol x}).$ In this extension, 
we propose to use a DNN to estimate the non-linear $f.$ Using multiple layers with non-linear activation, such as a rectified linear unit (ReLU), we can capture the non-linear relationship between the covariate and the mean of the log-transformed failure time in the AFT model. The DNN thus provides the model capacity for the non-linear $f.$ 
The proposed network architecture is not fixed and varies in numerical study and analysis of real datasets, as described in Supplementary Material B. The proposed network configuration is illustrated in Figure~\ref{fig:architecture}, with a sketch and example provided (details are in Supplementary Material B). A key aspect of this configuration is the use of shared parameters. Note that the loss requires paired data, and we consider $(\boldx_1, \boldx_2, y_1, y_2, \Delta_1, \Delta_2)$ as one input.    
Let $\sigma(\cdot)$ and $ {\bf W}_l $ be an activation vector function and weight matrix for the $l^{th}$ layer, respectively. Then $f({\boldsymbol x})$ is parameterized by $f_{\bf w}$ where 
$$ f_{\bf w} ( \boldx ) = \sigma_L ( {\bf W}_L (  \sigma( {\bf W}_{L-1} ( \cdots  \sigma({\bf W}_l ( \cdots  \sigma(  {\bf W}_1 { {\boldsymbol x} }  )  \cdots   )  ) \cdots   )  ) ) ),$$
and ${ \bf w } = ( {\bf W}_1, \ldots, {\bf W}_L).$
Here,  
$\sigma(\cdot)$ is a vector function, 
such as $ \sigma(\boldsymbol x) = ( \sigma ( x_1) , \ldots , \sigma(x_p))^{\top}.$ The activation function for the last layer, $\sigma_L(\cdot)$, is a linear function, the output of which is a one-dimensional real number.  
Temporarily, we drop the bias terms (intercepts) for the simplicity of the notation, which can easily be added in practice. For example, we let ${\boldsymbol h}^{(t-1)}$ be the $t-1$th layer, and $d_{t-1}$ is the number of nodes in this layer. The hidden nodes of the $t$-th layer is constructed by the following:
\begin{eqnarray*}
{\boldsymbol h}^{(t)}_k = \sigma \left(\sum_{k^{'}=1}^{d_{t-1}} {\boldsymbol h}^{(t-1)}_{k^{'}} w^{(t)}_{k^{'}, k} \right), k = 1, \ldots, d_t
\end{eqnarray*}
 Here, $\sigma(x)$ is the non-linear function, e.g., $x_{+} = \max \{ x, 0 \}$ (ReLU). 
The proposed network uses the pair-wise loss with respect to two inputs and weight-sharing for $f_{\bf w}( {\boldsymbol x}_1)$ and $f_{\bf w}( {\boldsymbol x}_2).$ 
We can construct a matched dataset by sub-sampling (see Section~\ref{sec:mod:subsamp}). 
After that, random samples of size $b$ are selected from the matched dataset, where $b$ is the size of the mini-batch.
In learning the Gehan loss defined in the equation~(\ref{wee_rank_geh}), an SGD algorithm (See Supplementary Material A for its theoretical properties) is used with a mini-batch. Detailed learning setups for the DNN, including architecture, batch size, epochs, and optimizers, are referred to in Supplementary Material B. 
In numerical studies, we use multi-layer perception (MLP), which has input, three or four hidden, and output layers. The real datasets have more complex architectures.    
Finally, we refer to our proposed DNN with a rank loss-based AFT model as `DeepR-AFT'. 

\begin{figure}
    \centering
    \begin{tabular}{cc}
    \includegraphics[width=120mm, height=70mm]{fig/Figure1-1-eps-converted-to.pdf}
    \includegraphics[width=50mm, height=70mm]{fig/Figure1-2-eps-converted-to.pdf}      
    \end{tabular}
    \caption{Example of architecture for DeepR-AFT. The outputs $f({\boldsymbol x}_1)$ and $f({\boldsymbol x}_2)$ share the parameters, and $\ell (f({\boldsymbol x}_1), f( {\boldsymbol x}_2), y_1, y_2, \Delta_1, \Delta_2 ) = \mathbb{I} ( \Delta_1=1) [ \log y_1 - \log y_2 - f( {\boldsymbol x}_1 ) + f({\boldsymbol x}_2) ]^{-}.$ The figures on the left and right illustrate the sketch of architecture and an example for Setup 3 in the Simulation section. One sample in the mini-batch appears as $(\boldx_1, \boldx_2, y_1, y_2, \Delta_1, \Delta_2).$} 
    \label{fig:architecture}
\end{figure}

\subsection{Sub-sampling pairs with DeepR-AFT} \label{sec:mod:subsamp}
The Gehan loss~\eqref{eqn:g2loss}, which requires a pair-wise evaluation of all pairs from the $n$ samples, has a computational complexity of $O(n^2)$ in a brute manner. As the Associate Editor suggested, this complexity can be significantly reduced to $O(n)$ by employing a simple sorting process. This adjustment allows the loss function to be rewritten as  
$ \sum_{i=1}^{n} \Delta_i \left[ \sum_{j=1}^{n} e_j( {\bf w} ) - (n-i+1) e_i({\bf w}) \right ] $ under the condition that $ e_1 ( {\bf w}) \ge \ldots \ge e_n ( {\bf w} ).$ 

Another strategy to decrease this computational complexity involves sub-sampling pairs rather than evaluating all possible pairs.
This sub-sampling idea has been explored recently in the context of completely observed massive data with large $n$ \citep{ma2014statistical,wang2018optimal,wang2021optimal}. For censored survival data, however, the literature is rather limited. Current sub-sampling methods are typically developed within the frameworks of the additive hazards model and Cox proportional hazard model with rare events, respectively.\citep{zuo2021sampling,  keret2020optimal}     
While the sorting process is also effective in reducing the computational complexity and seems promising, in this paper, we consider a {\rm simple} sub-sampling procedure mainly due to its relative simplicity in the tuning of the DNN.  

Since the Gehan loss is formed by constructing all possible pairs between all residuals and residuals concerning non-censored failure times, we propose a sub-sampling approach from all pairs in the Gehan loss, i.e., sampling from $\{ ( y_i, y_j ) \}_{  \Delta_i = 1, j \neq i }.$ For each non-censored failure time $y_i $, we sample $s$ pairs such that $(y_i, y_j)$ from $n-1$ possible pairs including the $y_i$. Finally, we obtain sub-samples from all possible pairs whose {\rm dimension is}  $ m = n^{'} \times s$ where $n^{'} = \sum_{i=1}^n \Delta_i.$ 
The proposed sub-sampling procedure has the following merits. First, it reduces the computational burden. 
Second, this sub-sampling procedure still works for the SGD algorithm because the resulting sample can be regarded as a random sample from all possible pairs in Gehan loss. A mini-batch for the SGD algorithm can be obtained through the $b$ sub-samples from the $ m $ sub-sampled pairs. The loss for each mini-batch with size $b$ is given by
\begin{eqnarray}
    \sum_{i=1}^k \sum_{j=1}^{r_i} \Delta_i \left[ e_i ( \boldw ) - e_j ( \boldw )   \right]^{-}, 
    \label{eqn:bloss}
\end{eqnarray}
where $\sum_{i=1}^k r_i = b$. Here, the $k$ is the number of non-censored failure times (indexed by $i$) in a mini-batch. 
For a comprehensive understanding, we provide a toy example of $n=6$. Consider the data $ (y_1, \Delta_1 =1, {\boldsymbol x}_1 ), (y_2, \Delta_2 = 1, {\boldsymbol x}_2 ), (y_3, \Delta_3 =1, {\boldsymbol x}_3 ), (y_4, \Delta_4 =0, {\boldsymbol x}_4 ), (y_5, \Delta_5 = 0, {\boldsymbol x}_5 ), (y_6, \Delta_6 =0 , {\boldsymbol x}_6 ).$ Set $s=2$ and $b=3$. Then, we can then have sub-paired samples such as 
$ \{ ( y_1, \Delta_1 =1, {\boldsymbol x}_1, y_4, \Delta_4 =0, {\boldsymbol x}_4 ), ( y_1, \Delta_1 =1, {\boldsymbol x}_1, y_2, \Delta_2 = 1, {\boldsymbol x}_2 ), ( y_2, \Delta_2 =1, {\boldsymbol x}_2, y_4, \Delta_4 =0, {\boldsymbol x}_4 ), ( y_2, \Delta_1 =1, {\boldsymbol x}_1, y_5, \Delta_5 = 0, {\boldsymbol x}_5 ), (y_3, \Delta_3 =1, {\boldsymbol x}_3, y_5, \Delta_5 = 0, {\boldsymbol x}_5  ), (y_3, \Delta_3 =1, {\boldsymbol x}_3, y_6, \Delta_6 = 0, {\boldsymbol x}_6  ) \},$ resulting in $m=6.$ In one epoch, a mini-batch can be $\{ ( y_1, \Delta_1 =1, {\boldsymbol x}_1, y_4, \Delta_4 =0, {\boldsymbol x}_4 ), ( y_1, \Delta_1 =1, {\boldsymbol x}_1, y_2, \Delta_2 =1, {\boldsymbol x}_2 ), (y_3, \Delta_3 =1, {\boldsymbol x}_3, y_5, \Delta_5 = 0, {\boldsymbol x}_5  ) \}$ with $b=3$, while other samples in sub-samples form another mini-batch. Two mini-batches can thus be used for an epoch. 

The size of $s$ does not seem to make a discernible difference in the prediction aspect. Results from simulation studies in this aspect are provided in Supplementary Material C.

{\noindent
{\bf Remark.} Our empirical observations validate the convergence or stability of loss across epochs. Refer to Supplementary Material B-3 for the discussion, including the requisite computational expenses, which may be improved by utilizing an advanced GPU unit and more sophisticated optimizers.    
}

\section{Simulation} \label{sec:sim}
\subsection{Setting} \label{sec:sim:set}
To evaluate the finite sample performance of the proposed DeepR-AFT, 
we conducted extensive simulation experiments. 
We assumed that, given covariates, the failure times $T_i$s were generated from an AFT model, $\log T_i \sim f( {\boldsymbol x}_i) + \epsilon_i, \ i =1, \ldots, n$. We considered three covariates, ${\boldsymbol x}_i = (x_{i1}, x_{i2}, x_{i3})^{\top}$ where $x_{i1} \sim Bernoulli(0.5)$, $x_{i2} \sim \mathcal{N} (x_{i1}/2, 1.0)$, and $x_{i3} \sim \mathcal{N} (x_{i2}/2,1.0)$. For the functional form of the mean part, $f(\boldx)$, we considered three forms of which two were non-linear and one was linear, specifically, (i) a non-linear function including an interaction term: $f({\boldsymbol x}) = 2 x_1 + x_2 x_3 +2 x_3$, (ii) a generalized additive model (GAM)-type function: $f( {\boldsymbol x} ) = x_1 + 0.5 x_2^2 + \exp(0.1 x_3)$, and (iii) a purely linear function: $f({\boldsymbol x}) = x_1 + 2x_2 + 2x_3$.
Four error distributions were evaluated: standard normal, Gumbel, $t$ with three degrees of freedom, and Laplace$(0, 1)$. The error distributions were shifted and scaled to have a zero mean and unit variance. Potential censoring times, $C_i$s, were generated from $\tau*U$ where $U$ is a uniform $(0, 1)$ distribution. The $\tau$ values were set to $20$, $40$ and $60$, which resulted in the censoring proportions of $0.42$, $0.34$, and $0.30$, respectively, for each simulation scenario. 

For each simulation setting, we generated training and test datasets independently. 
For the training dataset, we considered $1000$ and $5000$ as the sample sizes. 
The sample size for the test dataset was $2000$. 
The computational costs of evaluating the Gehan loss were too high, as it requires pair-wise comparisons. Instead, we used a mini-batch approach, which converges to the (local) minimum of the loss function using the full data. 
We sub-sampled pairs from the training data. 
In the evaluation of the Gehan loss, pairs are assessed for each uncensored survival time, which ensures that each pair includes at least one uncensored time. We implement a sub-sampling strategy at the pair level.
We used this sub-sampled pair as the training data pairs based on the properties of the SGD algorithm. 
In (iii), the linear model, we used linear activation for the setup. Using linear activation is similar to working with linear models. However, the nodes of each layer decrease in the applied DNN, having the side effect of a dimension reduction. This tends to give the DNN an advantage in large-dimensional data analysis. We also conducted numerical studies to observe this side effect in more detail in Section~\ref{sec:sim:hcov}. We then input covariates in the test dataset to calculate the predicted failure times. To compare the performances of the proposed DeepR-AFT, we also considered the existing estimators based on PAFT and SAFT models. For the PAFT model, the error distribution was assumed to follow a normal distribution with a constant variance. For the SAFT model, the error distribution was left unspecified. As measures for assessing prediction performance, we considered the following: the mean squared error (MSE) and the concordance index (C-index). MSE is defined as 
\begin{eqnarray}
    \text{MSE} = \frac{1}{n} \sum_{i=1}^n \left\{ \hat{f}( {\boldsymbol x}_i) - f( {\boldsymbol x}_i) \right\}^2.  
\label{eq:mse}
\end{eqnarray}
where $\hat{f}$ denotes an estimated $f$. The C-index is defined as
\begin{eqnarray}
 \frac{\sum_{i \neq j } \Delta_i   \mathbb{I} ( Y_i < Y_j ) \mathbb{I} ( \hat{f}( {\boldsymbol x}_i) < \hat{f} ({\boldsymbol x}_j)  )  }{ \sum_{i \neq j} \Delta_i  \mathbb{I}( Y_i < Y_j) }
\label{eq:cindex0}. 
\end{eqnarray}
In addition, the inverse-probability-of-censoring weighted (IPCW) C-index is defined as:
\begin{eqnarray}
\frac{\sum_{i \neq j } \Delta_i   \mathbb{I} ( Y_i < Y_j ) G(Y_i)^{-2} \mathbb{I} ( \hat{f}( {\boldsymbol x}_i) < \hat{f} ({\boldsymbol x}_j)  )  }{ \sum_{i \neq j} \Delta_i  G(Y_i)^{-2}  \mathbb{I}( Y_i < Y_j) }
\label{eq:cindex}
\end{eqnarray}
where $\hat{G}$ is the Kaplan-Meier estimator for the censoring distribution. It is known that IPCW C-index ensures the consistency to $\mathbb{P} ( \hat{f} (x_i) > \hat{f} (x_j) \mid T_i < T_j  )$  
\citep{kang2015comparing}. Therefore, we use the IPCW C-index denoted by `C-index' in this paper. 

The MSE is designed to capture differences in the estimated and true means in the log-transformed $T$, so it is a natural performance measure for the AFT model that models the mean of the log-transformed $T$. This measure has also been considered in assessing the prediction accuracy of AFT models \citep{ding2015estimating, seo2021accelerated}. 
The C-index is the most commonly used measure of prediction accuracy in survival analysis. \citep{Harr:Flem:clas:1982,katzman2018deepsurv,kvamme2019time}. It is the proportion of the pairs of the predicted failure times that maintain the correct order in the observed failure times. Unlike the MSE, the C-index is bounded between 0 and 1. The higher, the better, taking $0.5$ when the prediction is randomly performed. In addition, it is closely related to the area under the curve, another popular measure of prediction accuracy in survival analysis. {To revisit for clarification, the detailed setups for architectures, learning process, and sub-sampling are deferred to Supplementary Material B.}  
 
\subsection{Estimation of various $f$} \label{sec:sim:mse}    
We first investigated the accuracy of estimation by the MSE and C-index for our proposed DeepR-AFT under various forms for $f$. 
To compare the performances, we also considered the PAFT and SAFT. 
For estimation of $f$ under PAFT and SAFT, a linear predictor for $f$ was considered, i.e., $f(\boldx) = \boldx^{\top}\boldbeta$ whose estimation can be carried out by plugging in estimates for $\boldbeta$. For the estimation of $\boldbeta$, an MLE and induced smoothed rank-based estimator with the Gehan-type weight were applied for PAFT, SAFT, and Random Survival Forest (RSF), respectively. Neither the PAFT nor the SAFT is designed to address the non-linearity in $f.$ The RSF is a non-parametric method known to effectively capture non-linearity\citep{ishwaran2008random}. 
In the RSF, a mean function of $T$ cannot be directly estimated; instead, the K-M estimator from the RSF results is utilized to compute the mean.

The implementation of the DeepR-AFT was carried out using \texttt{python3 keras}, and the calculations were conducted using Titan-RTX graphics processing units. To calculate the estimates under PAFT, SAFT, we applied the \texttt{survreg} function in the \texttt{survival} package \citep{survival-package} and the \texttt{aftsrr} function in the \texttt{aftgee} package \citep{Chio:Kang:Yan:fitt:2014} in R \citep{r2021}. For the RSF, we applied the \texttt{rfsrc} function in the \texttt{randomForestSRC} package \cite{ishwaran2008random}.   

First, we report the simulation results for (i), $f({\boldsymbol x}) = 2 x_1 + x_2 x_3 +2 x_3$, having an interaction. Here, we consider an interaction term between $x_2$ and $x_3$ while the others are linear. A model with a relatively low signal of the main effects was used to investigate the interaction effect. The results are summarized in Table 1. The proposed DeepR-AFT model showed superior performance to the SAFT and PAFT in all cases in terms of the MSEs and C-indices. 
All the MSEs under the proposed DeepR-AFT were smaller than those under the PAFT and SAFT. The reductions in the MSEs were larger for a larger sample size with Gumbel and $t(3)$ errors, whose distributions were either skewed or heavy-tailed. The MSE under the proposed DeepR-AFT was approximately $0.45$ times those under the PAFT and SAFT. Similar patterns were found for the C-indices. The advantage of the proposed DeepR-AFT is that the DNN can capture the interaction effect in the mean function.
When compared to the RSF, DeepR-AFT demonstrates notable superiority in terms of MSE. Furthermore, DeepR-AFT's advantage is evident in the Gaussian and Gumbel C-index especially with large samples. Additionally, DeepR-AFT with respect to the C-index across different error distributions and censoring rates is more stable.

Second, Table 2 shows the results for (ii), $f({\boldsymbol x}) = x_1 + 0.5 x_2^2 + \exp(0.1 x_3)$, a GAM-type function that includes a non-linear relationship, but without any interactions.
The findings were similar to those under (i). The proposed DeepR-AFT produced similar or smaller MSEs and higher C-indices than those under the PAFT and SAFT. 
The proposed DeepR-AFT was superior to the PAFT or SAFT in terms of the MSE in all cases. 
It seems that the gain in the C-index can be larger than that in the MSE, because we use the rank-based Gehan loss, i.e., built on the ranks of residuals.
Compared to the RSF, DeepR-AFT performs better in the C-index for Gaussian and Gumbel distributions with large sample sizes. In terms of MSE, the RSF demonstrates slightly better performance in most scenarios. However, this gain in MSE is relatively minimal for Gaussian and Gumbel distributions. Additionally, DeepR-AFT is more robust concerning varying error distributions, whereas RSF exhibits greater fluctuation.

Third, the simulation results for (iii), $f({\boldsymbol x}) = x_1 + 2x_2 + 2x_3$, a simple linear relationship, are summarized in Table 3. We used a linear activation function in the proposed DeepR-AFT.
The results in Table 3 show that the overall performance of the non-DeepR-AFTs is better than those of the proposed DeepR-AFT in MSEs, especially for the smaller sample size ($n = 1000$). However, the gaps in the MSEs decreased as the sample size increased. Notice that DNNs are designed to capture trends that go beyond a simple linear relationship. Consequently, when the true relationship is linear, estimated $f$s under the proposed DeepR-AFT tend to have larger variability than those designed to capture just a simple linear trend. This may explain why the MSEs under the PAFT and SAFT were smaller than those under the DeepR-AFT when the underlying mean function was linear. 
Meanwhile, the C-indices of the proposed DeepR-AFT were either comparable to or higher than those of the PAFT. This is, again, mainly due to the nature of the Gehan-type loss function that uses the ranks of the residuals. In contrast to the previous cases, the RSF is markedly inferior to the other models in terms of MSE and C-index across all error distributions.

In summary, the proposed DeepR-AFT performed better than the PAFT and SAFT when the underlying mean functions were complex with interactions or had non-linear relationships and when the sample sizes were large. Under the mean function with a simple linear relationship, the DeepR-AFT produced mixed results. When the MSE is considered, the proposed DeepR-AFT appeared to be slightly inferior to the PAFT and SAFT. On the other hand, it showed comparable or better performances in terms of the C-index. {The DeepR-AFT demonstrates superior robustness and overall performance compared to the RSF, which exhibits superiority only in specific scenarios.}    

In Section~\ref{sec:sim:hcov}, we further investigate the performance of our proposed DeepR-AFT when the linear predictor is assumed for the mean part but with large-dimensional covariates, with only a few affecting the mean function. 


\begin{table}[tp]
\caption{Simulation results for $f({\boldsymbol x}) = 2 x_1 + x_2 x_3 +2 x_3$. Larger values of $\tau$ correspond to smaller censoring rates. Four error distributions are considered: standard normal (Gaussian), Gumbel (Gumbel), Laplace (Laplace), and $t$ with $3$ degrees of freedom ($t(3)$). Results are MSEs and C-indices in parentheses averaged over $100$ replicates with two sample sizes $n = 1000$ and $5000$ for training data sets.}
\centering
\label{tab:sim_set1}
\footnotesize
\begin{tabular}{ll|cccc|cccc}\hline
 & & \multicolumn{4}{c|}{$n = 5000$} & \multicolumn{4}{c}{$n = 1000$}  \\      &    $\tau$ &    DeepR-AFT & RSF &nPAFT & SAFT & DeepR-AFT & RSF & PAFT & SAFT  \\\hline
Gaussian & 20  & 1.502 (0.887)   &  3.732 (0.885)  & 2.294 (0.862) & 2.108 (0.864) & 
1.514 (0.872)  &    3.656 (0.881)	     & 2.353 (0.860) & 2.154 (0.862) \\ 
 & 40  &  1.212 (0.885) &   3.262 (0.883)   & 2.164 (0.861) & 2.000 (0.862) & 
 1.274 (0.871)  &  3.091 (0.871)   & 2.165 (0.860) & 2.009 (0.862) \\ 
& 60  &  1.060 (0.885)  &  2.836 (0.883)    &  2.099 (0.860) & 1.953 (0.861) & 
1.172 (0.871) & 	 3.413 (0.879)   & 2.075 (0.860) & 1.930 (0.861)  \\\hline 
Gumbel & 20 &  0.959 (0.904) & 	4.296 (0.891)
   & 2.353 (0.866) & 2.154 (0.869) & 
   1.600 (0.891)  & 4.650 (0.891)   & 2.356 (0.866) & 2.137 (0.869)\\
& 40  &   0.884 (0.902) &  3.231 (0.891)  & 2.171 (0.864) & 1.978 (0.866) &  
1.314 (0.888)  & 3.274 (0.883) & 2.190 (0.864) & 1.992 (0.866) \\
& 60  &  0.726 (0.900)   &  2.968 (0.891)   & 2.101 (0.864) & 1.923 (0.866) 
&  1.203 (0.888)  &  2.306 (0.874) &  2.100 (0.863) &  1.920 (0.865) \\  \hline  
Laplace & 20 &  1.203 (0.887)  & 	4.056 (0.900)  & 2.290 (0.867) & 2.127 (0.869) & 1.526 (0.872) & 4.745 (0.881) &  2.312 (0.866) & 2.137 (0.868) \\ 
 & 40 &   1.010 (0.886)  &  3.040 (0.891) & 2.157 (0.865) & 2.006 (0.867) & 
 1.328 (0.871)  &  	3.435 (0.875) & 2.166 (0.865) & 2.024 (0.867)   \\
& 60 &   0.940 (0.886)  &  2.740 (0.876)  & 2.060 (0.866) & 1.924 (0.867) &  
1.544 (0.870)  &  3.003 (0.874)    & 2.110 (0.864) & 1.974 (0.866)  \\ \hline 
 $t(3)$ & 20 & {1.602 (0.894)}  & 4.671 (0.910)  & 2.317 (0.875) & 2.178 (0.877) &
 1.879 (0.890) & 4.221 (0.901) & 2.330 (0.875) & 2.188 (0.877)   \\
& 40 & 1.551 (0.893)   & 3.145 (0.898)  & 2.161 (0.874) & 2.025 (0.875) &  
1.730 (0.890) & 3.932 (0.898) & 2.144 (0.873) & 2.010 (0.874) \\
& 60 &  1.045 (0.895)  &  2.773 (0.889)  & 2.053 (0.873) & 1.930 (0.875) & 
1.418 (0.890)  & 2.481 (0.890) & 2.064 (0.872) & 1.944 (0.874)
\\\hline 
\end{tabular} 
\end{table}

       
\begin{table}[htp] 
\caption{Simulation results for $f({\boldsymbol x}) = x_1 + 0.5 x_2^2 + \exp(0.1 x_3)$. Larger values of $\tau$ correspond to smaller censoring rates. Four error distributions are considered: standard normal (Gaussian), Gumbel (Gumbel), Laplace (Laplace,) and $t$ with $3$ degrees of freedom ($t(3)$). Results are MSEs and C-indices in parentheses averaged over $100$ replicates with two sample sizes $n = 1000$ and $5000$ for training data sets.}
\label{tab:sim_set2}
\centering
\footnotesize
\begin{tabular}{ll|cccc|cccc}\hline
 & & \multicolumn{4}{c|}{$n = 5000$ } & \multicolumn{4}{c}{$n = 1000$}  \\      &    $\tau$ &    DeepR-AFT & RSF &  PAFT & SAFT & DeepR-AFT & RSF &  PAFT & SAFT  \\\hline 
Gaussian & 20  & 0.291 (0.718)  & 0.328 (0.701)
 &  0.589 (0.660) & {  0.577} (0.660) & 0.354 (0.708) &  0.306 (0.690)
  & 0.596 (0.661) & {  0.586} (0.661)\\
& 40  &  0.204 (0.720) & 0.216 (0.713)  & 0.581 (0.660) & 0.572 (0.661) & 0.252 (0.711)  &  0.179 (0.709) & 0.589 (0.660) & 0.580 (0.661) \\
 & 60  &  0.171 (0.719)  & 0.175 (0.716)   & 0.581 (0.659) & 0.574 (0.659) &  0.216 (0.711)  &  0.238 (0.712)  & 0.578 (0.658) & 0.572 (0.659))\\
 \hline
Gumbel & 20 &  0.370 (0.754)  & 0.331 (0.720)
  &  0.620 (0.672) & 0.589 (0.672) & 0.422 (
0.722)  & 0.424 (0.725)
   &  0.611 (0.670) &  0.582 (0.671)\\ 
& 40  &  0.279 (0.744)  &  0.252 (0.735)   & 0.591 (0.669) & 0.572 (0.669) & 0.355 (0.719)  & 0.259 (0.731)   & 0.599 (0.669) & 0.581 (0.670) \\
& 60  & 0.256 (0.737)  & 0.137 (0.729)  & 0.587 (0.671) & 0.573 (0.671) &  0.328
(0.717)  & 
0.179 (0.722)   & 0.587 (0.668) &  0.573 (0.669) \\
\hline
Laplace & 20 & 0.287 (0.718)   &   0.316 (0.736)
 & 0.586 (0.675) & 0.580 (0.675) &  
  0.394 (0.706)   & 0.465 (0.742)
& 0.600 (0.673) & {  0.596} (0.674) \\ 
& 40 &   0.211 (0.723)  &  0.190 (0.738)
 & 0.579 (0.674) & 0.570 (0.674) & 0.307 (0.709)   &  0.179 (0.731)   & 0.591 (0.673) & 0.583 (0.673) \\
& 60 & 0.177 (0.727)  &  0.191 (0.719)  & 0.577 (0.672) & 0.570 (0.672) &  0.277 (0.710)  & 0.151 (0.734)   & 0.589 (0.672) & 0.581 (0.673) \\
\hline
$t(3)$ & 20 &  0.424 (0.720)  & 0.359 (0.760)
  & 0.592 (0.689) & 0.589 (0.689) & 0.422 (0.722)   &  0.347 (0.752)
&  0.594 (0.687) & {  0.592} (0.688) \\ 
& 40 & 
0.364 (0.725)  & 0.204 (0.755)  & 0.577 (0.686) & 0.570 (0.686) &  0.355 (0.719)  & 0.273 (0.764) & 0.582 (0.686) & 0.574 (0.686) \\
& 60 &  0.392 (0.721) &  0.143 (0.750)   & 0.578 (0.687) & 0.571 (0.687) &  0.328 (0.717) & 0.147 (0.755)  & 0.583 (0.684) & 0.576 (0.685) \\
\hline 
\end{tabular}
\end{table}


\begin{table}[htp] 
\caption{Simulation results for $f({\boldsymbol x}) = x_1 + 2x_2 + 2x_3$. Larger values of $\tau$ correspond to smaller censoring rates. Four error distributions are considered: standard normal (Gaussian), Gumbel (Gumbel), Laplace (Laplace), and $t$ with $3$ degrees of freedom ($t(3)$). Results are MSEs and C-indices in parentheses averaged over $100$ replicates with two sample sizes $n = 1000$ and $5000$ for training data sets.}
\centering
\label{tab:sim_set3}
\footnotesize
\begin{tabular}{ll|rccc|rccc}\hline
& & \multicolumn{4}{c|}{$n = 5000$} & \multicolumn{4}{c}{$n = 1000$}  \\      &    $\tau$ &    DeepR-AFT & RSF & PAFT & SAFT & DeepR-AFT & RSF & PAFT & SAFT  \\\hline
Gaussian & 20  & 0.034 (0.924) & 3.352 (0.912)
   & { 0.002} (0.929) & 0.002 (0.929) & 0.128 (0.929)  &   3.358 (0.910)
 & { 0.008} (0.928) & 0.008 (0.928)\\
& 40  & 0.040 (0.930)  &  2.575 (0.908) & { 0.002} (0.926) & 0.002 (0.926) &  0.127 (0.926)  &
2.828 (0.898)
 & { 0.007} (0.925) & 0.007 (0.925) \\
& 60  &  0.043 (0.924) &  2.037 (0.904)  & { 0.001} (0.924) & 0.001 (0.924) &   0.094 (0.925) &  2.615 (0.905)  & { 0.007} (0.924) & 0.007 (0.924) \\
\hline
 Gumbel & 20 &   0.101 (0.943)  & 3.678	(0.924)
 & 0.001 (0.934) & { 0.001} (0.934) &   1.056 (0.943)  &  4.055 (0.922)
 & 0.008 (0.934) &  { 0.007} (0.934) \\ 
& 40  &   0.090 (0.940)  &  2.731 (0.918) & 0.001 (0.931) & { 0.001} (0.931) &  
0.531 (0.940)  &   2.840 (0.911) & 0.007 (0.930) & { 0.006} (0.930) \\
& 60  &    0.080 (0.939) &  
2.255 (0.913)  & 0.001 (0.929) & { 0.001} (0.929) &   0.461 (0.938)  &  2.092 (0.898)  & 0.005 (0.928) & { 0.005} (0.928) \\
\hline 
Laplace & 20 &  0.036 (0.933)  & 3.463 (0.925)
 & 0.002 (0.934) & { 0.001} (0.934) & 0.165 (0.933)  & 3.955 (0.917)
 & 0.008 (0.934) & { 0.006} (0.934) \\ 
& 40 &  0.040 (0.930)  & 2.469 (0.912)  & 0.001 (0.931) & { 0.001} (0.931) &  0.194 (0.930)  & 2.820 (0.906)  & 0.007 (0.930) & { 0.005} (0.931) \\
& 60 &    0.041 (0.929) & 2.142 (0.904)   & 0.001 (0.929) & { 0.001} (0.929) &  0.121 (0.929) &  2.437 (0.901)  & 0.006 (0.929) & {0.005} (0.929) \\
\hline 
$t(3)$ & 20 &  0.111 (0.940)  & 3.771 (0.933)
 & 0.004 (0.941) & { 0.001} (0.941) & 0.383 (0.940)   &  3.634 (0.921)
& 0.010 (0.941) & { 0.005} (0.941) \\ 
& 40 &  0.100 (0.940) &  2.537 (0.921)
  & 0.002 (0.939) & { 0.001} (0.939) &  0.281 (0.938)   &  3.151 (0.913)
  & 0.009 (0.938) & { 0.005} (0.938) \\
& 60 &   0.096 (0.937)  & 2.071 (0.915)    & 0.002 (0.937) & { 0.001} (0.937) & 0.220 (0.937)  &  
2.149 (0.909)   & 0.007 (0.937) & { 0.004} (0.937) \\
\hline
\end{tabular}
\end{table}


\subsection{Simulation for biases and variance} \label{sec:sim:bias}   

In this subsection, we {investigate the bias and variance trade-off of the proposed DeepR-AFT.} 
Since an MSE can be decomposed into biases and variances, 
we evaluated the proportion of the bias and variance in each MSE.  
Using the learned $\hat{f}$s from the training sets, we examine the bias and variance on the newly generated samples. Among the setups considered in Section~\ref{sec:sim:mse}, we chose $\tau = 40,$ a training sample size of $3000$, and a Gaussian error distribution to obtain $\hat{f}_{k}s, \ k = 1, \ldots, 100$ with the non-linear interaction, GAM-type, and purely linear types. 
Note that the bias and variance must be evaluated at a true fixed $f(\boldx)$. 
The corresponding simulation procedures are summarized below:

\begin{enumerate}
    \item Generate a ${\boldsymbol x}_i = (x_{i1},x_{i2}, x_{i3})^{\top}$. 
    \item Obtain $100$ $\hat{f}_{ik} = \hat{f}_k( \boldsymbol x_i)$ for DeepR-AFT, PAFT, and SAFT where $\hat{f}_k(\cdot)$ is the learned $\hat{f}$ obtained from the $k$th training set of size $3000 \  (k=1, \ldots, 100)$   
   \item Calculate the squared bias, and variance based on $100$ $\hat{f}_{ik} $s.
    \item Repeat 1--4 from $i=1$ to $2000$.
\end{enumerate}

The squared bias and variance were measured for generated $\boldsymbol x_i,$ and the average and standard deviation were calculated for over $100$ $\hat{f}_{ik}$s.
The results are shown in Table~\ref{tab:bias-variance} (since the RSF has a severe bias, with the exception of GAM-type, we consider the comparison with SAFT and PAFT).       
The results showed that for the GAM-type and non-linear with an interaction, biases of the SAFT and PAFT were larger than those of the DeepR-AFT. In the linear case, however, the relationship was reversed; the DeepR-AFT had a larger bias. The DeepR-AFT model showed relatively higher variances with respect to the squared biases compared to the non-linear $f$s. Note that for the GAM-type and non-linear cases, the proportion of the squared biases was dominant for a given MSE, with at least $5$ times being larger than the variance. In the GAM-type and non-linear cases, considering larger discrepancies in biases and smaller MSEs of the DeepR-AFT, the variances have little effect on the DeepR-AFT performance, implying that the DeepR-AFT can address the bias well with small variance. In the linear relationship situation, as expected, the SAFT and PAFT showed very low squared biases with higher values of variances. The DeepR-AFT showed similar magnitudes of squared biases and variances; both were larger than those of the PAFT and SAFT, indicating an inferior performance. 

To investigate further the behavior of biases, we drew plots of the true $f({\boldsymbol x})$s versus averaged predicted $f({\boldsymbol x})$s (in red) overlaid by the bars representing the $0.025$ and $0.975$ quantiles (in black) based on $100$ replications (Figures 2-4).  
When the curve in each plot is closer to the unit slope line,  
it implies small biases. The plots showed that, for the linear case, all three methods seem to be virtually unbiased. Some biases were observed in the other two cases.  
The PAFT (Figure 4) and SAFT (Figure 3) displayed highly undulating patterns in both cases compared to those of the DeepR-AFT (Figure 2). For the GAM-type, the averaged predicted $f({\boldsymbol x})$s formed a pattern farther away from the unit slope line, which also explains the large degradation of the C-indices in Table 2. In summary, with respect to the MSEs, the DeepR-AFT provided better estimators for the mean functions with complex non-linear relationships, whereas its performance was not necessarily as good as those of the PAFT and SAFT when a simple linear relationship was postulated.
 
\begin{table}[tp]
    \centering
    \caption{Squared Biases and Variances for various $f$s. Results are squared biases (Bias$^2$), and variances (Var) averaged with the generating sample size $n = 2000.$} 
    \label{tab:bias-variance}    
     \begin{tabular}{c|c|c | c | c}\hline
 Model & Measure &  Linear & GAM-type & Non-linear \\\hline   
  \multirow{2}{*}{DeepR-AFT} & $\textrm{Bias}^2$  &  0.0306 (0.0439)  & 0.2906 (0.7328)   & 1.4929 (5.9032)   \\
   & $\textrm{Var}$ & 0.0312 (0.0369)  & 0.0588 (0.0374)   & 0.1278 (0.1629)  \\\cline{1-5}
      \multirow{2}{*}{SAFT}  & $\textrm{Bias}^2$ &   $<0.0001$ ($<0.0001$)  & 0.5687 (2.0995)   & 2.0783 (6.8243)   \\   
       &  $\textrm{Var}$ &  0.0025 (0.0015)  & 0.0022 (0.0013)   & 0.0041 (0.0027)    \\\cline{1-5}
      \multirow{2}{*}{PAFT} & $\textrm{Bias}^2$   &  $ < 0.0001$  ($<0.0001$)  & 0.5749 (2.0867)   & 2.2368 (7.3661)        \\
     &  $\textrm{Var}$  & 0.0024 (0.0015)  & 0.0021 (0.0012)   & 0.0044 (0.0031)    \\\cline{1-5}
    \end{tabular}
\end{table}    

\subsection{Large dimensional covariates} \label{sec:sim:hcov}    

In this subsection, we evaluate the performances of DeepR-AFT with large dimensional covariates when $f$ is linear. As shown in Section~\ref{sec:sim:mse}, in terms of the MSEs, the DeepR-AFT did not seem to perform better than the conventional AFT estimators. However, it has been reported that the DNN generally performs better with larger dimensional data \citep{hao2021deep}. Therefore, we conducted another simulation study concerning larger dimensional data. Specifically, we considered the setting with the linear predictor for $f$, i.e., $f({\boldsymbol x}) = x_1 + 2 x_2 + 2x_3$, using a training sample size of $1000$ and a Gaussian error. We added $K$-dimensional covariates with no effects generated from the standard normal distributions. The corresponding $f$ was then $f({\boldsymbol x}) = x_1 + 2 x_2 + 2x_3 + \sum_{k=4}^{K} \beta_{k}x_k$ where $\beta_k = 0$ for $4, \ldots, K$. For the number of added covariates, we considered $0$ to $1000$. 

Figure 5 shows the C-indices and $\log(MSE)$s evaluated in the test dataset of size $2000$ for different numbers of added covariates. As shown in Figure 5, the C-index for each model decreased as the dimensions increased. The decrease rate for the DeepR-AFT
was the smallest. Adding non-significant covariates did not seem to have a substantial effect in terms of the C-index.  
The DeepR-AFT still worked well when the dimension of the non-significant additive covariates was $1000$ with the corresponding C-index being $0.88$. However, this was not the case for the conventional AFT estimators under the PAFT and SAFT. 
Note that we only reported the results for the PAFT and SAFT when the numbers of the added non-significant covariates were $700$ or less and $300$ or less, respectively, due to computational limits.

\begin{figure}[bp]
    \centering
    \begin{tabular}{ccc}
    \includegraphics[width=35mm, height=35mm]{fig/Figure2-1-eps-converted-to.pdf}     &
       \includegraphics[width=35mm, height=35mm]{fig/Figure2-2-eps-converted-to.pdf}   &      
       \includegraphics[width=35mm, height=35mm]{fig/Figure2-3-eps-converted-to.pdf}   
        \end{tabular}
    \caption{Left to right: plots of predicted means based on the proposed DeepR-AFT versus true means for the linear (L), GAM-type (GAM), and non-linear including an interaction (NL) where the upper and lower bars denote the $0.025$ and $0.975$ quantiles, respectively.}
    \label{fig:hdm_dnn}
\end{figure}

\begin{figure}[tp]
    \centering
    \begin{tabular}{ccc}
    \includegraphics[width=35mm, height=35mm]{fig/Figure3-1-eps-converted-to.pdf}     &
       \includegraphics[width=35mm, height=35mm]{fig/Figure3-2-eps-converted-to.pdf}   &      
       \includegraphics[width=35mm, height=35mm]{fig/Figure3-3-eps-converted-to.pdf}   
        \end{tabular}
    \caption{Left to right: plots of predicted means based on the semiparametric AFT model versus true means for the linear (L), GAM-type (GAM), and non-linear including an interaction (NL) where the upper and lower bars denote the $0.025$ and $0.975$ quantiles, respectively.}
    \label{fig:hdm_saft}
\end{figure}

\begin{figure}[tp]
    \centering
    \begin{tabular}{ccc}
    \includegraphics[width=35mm, height=35mm]{fig/Figure4-1-eps-converted-to.pdf}     &
       \includegraphics[width=35mm, height=35mm]{fig/Figure4-2-eps-converted-to.pdf}   &      
       \includegraphics[width=35mm, height=35mm]{fig/Figure4-3-eps-converted-to.pdf}   
        \end{tabular}
    \caption{Left to right: plots of predicted means based on the parametric AFT model versus true means for the linear (L), GAM-type (GAM), and non-linear including an interaction (NL) where the upper and lower bars denote the $0.025$ and $0.975$ quantiles, respectively.}
    \label{fig:hdm_paft}
\end{figure}

The PAFT showed a drastic decrease when the dimensions exceeded $500$. The SAFT showed a similar pattern to that of the DeepR-AFT with dimensions being $300$ or less, but no results are available beyond this. This also shows the computational limitations of the conventional AFT models.   
The overall results for the MSEs were comparable to those for the C-indices, with slightly larger values for the DeepR-AFT than for the PAFT when the same number of additional non-significant covariates was considered. In this case, the MSEs of PAFT and AFT increased sharply compared with those of the DeepR-AFT as the dimension increased.   

In summary, when the dimensions of the added covariates are relatively small, the DeepR-AFT can preserve the order of the mean failure times in close proximity to their true order, despite the MSEs being generally higher. DeepR-AFT is superior to the other models considered here in terms of the MSE, C-index, and computational aspects in larger dimensions. This implies that, despite the application of linear activation, DNNs can still extract useful information from large dimensional inputs.

\begin{figure}    \centering
\begin{tabular}{ccc}
 \includegraphics[width=35mm, height=50mm]{fig/Figure5-1-eps-converted-to.pdf} & 
 \includegraphics[width=35mm, height=50mm]{fig/Figure5-2-eps-converted-to.pdf} & 
  \includegraphics[width=35mm, height=50mm]{fig/Figure5-3-eps-converted-to.pdf} \\
  \includegraphics[width=35mm, height=50mm]{fig/Figure5-4-eps-converted-to.pdf}  & 
  \includegraphics[width=35mm, height=50mm]{fig/Figure5-5-eps-converted-to.pdf}  & 
  \includegraphics[width=35mm, height=50mm]{fig/Figure5-6-eps-converted-to.pdf}     
\end{tabular}
\caption{Top to bottom: Plots of $\log$(MSE)s and C-indices and for different numbers of added non-significant covariates. Left to right: models considered - DeepR-AFT, PAFT, and SAFT.}
 \label{fig:HRS-D}
\end{figure}



\section{Real data analysis} \label{sec:data}

\subsection{Data descriptions}

In this section, we illustrate an application of our proposed DeepR-AFT by analyzing three datasets, referred to as {\it NB-SEQ}, {\it FL-CHAIN}, and {\it SEER}. 

The first dataset, {\it NB-SEQ}, comprises neuroblastoma data and other supervised penalty learning benchmarks (two neuroblastoma datasets and several CHIP-seq datasets; https://github.com/tdhock/neuroblastoma-data/blob/master/README.org). \citep{Rigaill2013, Hocking2017}. These datasets involve censored regression for supervised penalty function learning for changing-point detection. The dataset includes various summary statistics and the minimum and maximum values of $\lambda$, the hyper-parameter for the penalty function. We considered $4778$ sequence data with $24$ covariates listed in Table 3 in Supplementary Material D. The main response variable was the maximum value of estimated $\lambda$. Note that the maximum value of $\lambda$ cannot be observed in some cases; we only know that the value is greater than a certain value and, thus, the true value is censored. Usually, the optimal value of $\lambda$ can be related to various summary statistics with a non-linear relationship. Hence, we analyzed the value of $\lambda$ via survival analysis, assuming an AFT model. The censoring rate was $31$\%. Half of the data were used for training and the other half for testing. 

As the second dataset, we considered a subset of breast cancer data from the Surveillance, Epidemiology, and End Results ({\it SEER}) program. The subset was defined as patients who were diagnosed as having breast cancer in $1992$. The event of interest was death due to breast cancer. The follow-up proceeded until the end of 2007; those who did not develop breast cancer by the end of $2007$ were treated as censored. About $56$\% of patients were right-censored. We considered $19$ covariates, a list of which is provided in Table 3, in Supplementary Material D. 

{\it FL-CHAIN} \citep{dispenzieri2012use,kyle2006prevalence}, the third dataset we considered, is from a study that evaluated the relationship between serum-free light chain (FLC) and mortality risk. The dataset comprises $7874$ subjects, 
an age- and a gender-stratified random sample of the original $15759$ subjects. 
Death is the event of interest, while survival time is defined as the number of days between enrolment and death. The eight covariates in the dataset include age at baseline (in years), sex (female or male), the calendar year that the blood sample was obtained, the kappa portion of serum FLC, the lambda portion of serum FLC, the FLC group indicator, serum creatinine level, and a binary indicator for the history of a diagnosis of monoclonal gammopathy. 
After eliminating the samples of individuals with $0$ survival times and missing serum creatinine values, the censoring rate for the remaining $6521$ participants was $70$. 




\begin{table}[tp]
\centering
\begin{tabular}{| c|c | c | c | }\hline
Model & NB-SEQ  & SEER & FL-CHAIN  \\\hline
DeepR-AFT & 0.7317 & 0.8507  &  0.7564  \\
RSF  & 0.7370  & 0.7286   &  0.7541   \\
PAFT & 0.6604   &  0.7230    &  0.7982  \\ 
SAFT &  0.6558   &  0.7298 &  0.7999  \\  \hline
\end{tabular}
    \caption{The C-indices for the three real data sets: {\it NB-SEQ, SEER,} and {\it FL-CHAIN.}} 
    \label{tab:real-res}
\end{table}
{
In the architectures for the real datasets, the Conv1D\cite{keras1d} layers were considered to increase the strength except for the {\it FL-CHAIN}, which is well-known to perform well under the parametric fitting (detailed architectures and learning setups are deferred to Supplementary Material B). In {\it NB-SEQ, SEER}, and {\it FL-CHAIN}, around 60$\sim$66\% and 33$\sim$40\% (2873/1913, 5721/3814,  4347/2173) are used for training and test datasets, respectively.}  


\subsection{Results} 

\begin{figure}[bp]
\centering
\begin{tabular}{ccccc}
\includegraphics[width=30mm, height=30mm]{fig/Figure6-1-eps-converted-to.pdf} & 
\includegraphics[width=30mm, height=30mm]{fig/Figure6-2-eps-converted-to.pdf} & 
\includegraphics[width=30mm, height=30mm]{fig/Figure6-3-eps-converted-to.pdf} &
\includegraphics[width=30mm, height=30mm]{fig/Figure6-4-eps-converted-to.pdf} &
\includegraphics[width=30mm, height=30mm]{fig/Figure6-5-eps-converted-to.pdf}  \\
\end{tabular}
\caption{The PDPs of DeepR-AFT in {\it NB-SEQ} data.}
\label{fig5-1}
\end{figure}

\begin{figure}[tp]
\centering
\begin{tabular}{ccccc}
\includegraphics[width=30mm, height=30mm]{fig/Figure7-1-eps-converted-to.pdf} & 
\includegraphics[width=30mm, height=30mm]{fig/Figure7-2-eps-converted-to.pdf} & 
\includegraphics[width=30mm, height=30mm]{fig/Figure7-3-eps-converted-to.pdf} &
\includegraphics[width=30mm, height=30mm]{fig/Figure7-4-eps-converted-to.pdf} &
\includegraphics[width=30mm, height=30mm]{fig/Figure7-5-eps-converted-to.pdf}  \\
\end{tabular}
\caption{The PDPs of DeepR-AFT in {\it SEER} data.}
\label{fig5-2}
\end{figure}

\begin{figure}[tp]
\centering
\begin{tabular}{ccccc}
\includegraphics[width=45mm, height=30mm]{fig/Figure8-1-eps-converted-to.pdf} & 
\includegraphics[width=45mm, height=30mm]{fig/Figure8-2-eps-converted-to.pdf} & 
\includegraphics[width=45mm, height=30mm]{fig/Figure8-3-eps-converted-to.pdf} \\
\end{tabular}
\caption{The PDPs of DeepR-AFT in {\it FL-CHAIN} data.}
\label{fig5-3}
\end{figure}


The prediction performance was measured by the C-index evaluated in the test data. The PAFT and SAFT were also considered, and we additionally considered the RSF, which was shown to perform well in the presence of a non-linearity in the simulation experiments. 
Note that the MSE could not be evaluated. The results are presented in Table 5.  
The results showed that for {\it NB-SEQ} and {\it SEER}, DeepR-AFT performed the best. non-linearity in the mean function was obvious since DeepR-AFT and RSF performed much better than the PAFT and SAFT. Also, the superiority of DeepR-AFT to RSF implies that the normality of the error term assumed in RSF is not likely to be satisfied in {\it SEER}. In {\it NB-SEQ,} the performance of RSF and DeepR-AFT are similar.  
For {\it FL-CHAIN}, all four methods performed similarly. This implies that the true mean function is likely to be linear. 
DeepR-AFT performed slightly worse, but the discrepancies were negligible.
  
\subsection{Sensitivity analysis}

In this subsection, we examine the effects of covariates on the mean function of the AFT model. Among various approaches, the partial dependence plot (PDP)\cite{friedman2001greedy} is considered due to its wide use in machine learning literature and easy applicability to DNNs. The PDP for a covariate shows its marginal effect. 
Since the number of covariates in each dataset was large, we provided the results for all covariates in Supplementary Material D. This section presents the five ({\it NB-SEQ} and {\it SEER}) or three ({\it FL-CHAIN}) PDPs showing the relatively large deviations.  

Figure 6 implies that in {\it NB-SEQ}, the increasing values of standard deviations in $\log (x)$ ($log.sd$) {or $\log(1+x)$ ($log.1.sd$)} has the effect of longer mean failure times in a fairly increasing fashion. The 25$^{th}$ percentile ($quartile.25$) had the opposite effect {while maintaining an almost monotone trend.} However, the 100$^{th}$ percentile in the log-log scale, i.e., maximum ($log.log.quartile.100$) is not simple to interpret. It showed a decreasing effect on negative values and an upward effect on positive values.
The maximum in the $\log (1+x)$ scale ($log.1.quartile.100$) had a quadratic effect and revealed that larger values of the maximum can have an 
 quadratic effect on the mean. 
In {\it SEER} (Figure 7), the effects of covariates were somewhat complicated except for {\it Diagnostic$\_$confirmation}. This non-linearity for each covariate could lead to larger values of the C-index when we use DeepR-AFT imposing non-linear mean functions for the AFT model. For example, Histologic,Type.ICD.O.3 had a skewed and V shape, i.e., the mean failure time reached its minimum in the left-value of this covariate, whereas the mean failure time is minimized around the largest value of {\it Laterality}. Finally, {\it FL-CHAIN} in Figure 8 shows a relatively simple shape in PDPs, a linear effect; this is consistent with the observation that the AFT models imposing a linear function have larger C-index values.
   


\section{Discussion} \label{sec:disc}

We studied AFT models combined with a DNN. Extensive simulation experiments and application to real datasets demonstrated that the DNN can effectively handle complex non-linear-nonparametric AFT models. Sub-sampling loss with the SGD is effective in achieving the gain of the DNN. The DeepR-AFT model proposed here can achieve a higher C-index in a linear mean model, although the MSE is not superior to conventional AFT models. Non-linear mean AFT models exhibit the relatively robust gains of DeepR-AFT, as measured by the C-index and MSE, whereas the RSF shows some sensitivity to error distribution through good performance.   

The applications to real datasets show that non-linear modeling of the mean function is recommended unless a linear trend in the mean function is obvious. A sensitivity analysis using PDPs also confirmed the existence of non-linear relationships between some covariates and the mean function, especially for {\it NB-SEQ} and {\it SEER}.
Additionally, the simulation experiments with large dimensional covariates demonstrated that the DeepR-AFT can extract useful information for prediction, despite the existence of many noisy covariates. 

Some future research directions are as follows. A more automated algorithm for tuning the learning rate and other hyper-parameters would be worthy of investigation. The current sub-sampling procedure uses simple random sampling. Applications of more optimal sub-sampling procedures could lead to better performance in terms of prediction accuracy.  
Clustered failure time data are frequently encountered in biomedical studies. Due to the correlated nature of clustered failure times, our proposed DNN method based on the independence assumption cannot be applied directly. However, the Gehan loss considered with our model could be extended to accommodate clustered failure time data. Specifically, Gehan loss based on a marginal model approach with the working independence assumption \cite{jin2006rank} is a promising candidate as the loss function. Implementing the DNN and a more optimal sub-sampling procedure accounting for the clustered nature increases complexity, which is a natural direction for future research. Furthermore, the sorting process in optimization mentioned in Section 3.3 seems promising and valuable and could be further explored in conjunction with DNN.

 \section*{Acknowledgement}

This work was supported by the National Research Foundation of Korea(NRF) grant funded by the Korea government(MSIT) (RS-2023-00218377, RS-2024-00341883).

\section*{Data availability statement}

The data that support the findings of this study are openly available ({\it NB-SEQ} and {\it FL-CHAIN}) or can be obtained upon request from https://seer.cancer.gov/ for {\it SEER}.

\section*{Conflict of interest}
 The authors declare that they have no conflict of interest.



\bibliography{deep_aft}

\end{document}